\documentclass[letterpaper]{article} 
\usepackage[]{aaai_preprint}  
\usepackage{times}  
\usepackage{helvet}  
\usepackage{courier}  
\usepackage[hyphens]{url}  
\usepackage{graphicx} 
\usepackage{natbib}  
\usepackage{caption} 
\usepackage{algorithm}
\usepackage{algorithmic}

\usepackage{newfloat}
\usepackage{listings}
\DeclareCaptionStyle{ruled}{labelfont=normalfont,labelsep=colon,strut=off} 
\floatstyle{ruled}
\newfloat{listing}{tb}{lst}{}
\floatname{listing}{Listing}
\usepackage{amsmath, amsfonts}
\usepackage{booktabs}

\title{Mixture Manifold Networks: A Computationally Efficient Baseline for Inverse Modeling}
\author{
    Gregory P. Spell, \textsuperscript{\rm 1}
    Simiao Ren, \textsuperscript{\rm 1}
    Leslie M. Collins, \textsuperscript{\rm 1}
    Jordan M. Malof \textsuperscript{\rm 2}
}
\affiliations{
    \textsuperscript{\rm 1}Duke University, Department of Electrical \& Computer Engineering, Durham, NC USA\\
    \textsuperscript{\rm 2}University of Montana, Department of Computer Science, Missoula, MT USA\\

    \{gregory.spell, simiao.ren, leslie.collins\}@duke.edu, \ jordan.malof@umontana.edu
}

\usepackage{bibentry}

\begin{document}

\maketitle
\begin{abstract}
We propose and show the efficacy of a new method to address generic inverse problems. Inverse modeling is the task whereby one seeks to determine the control parameters of a natural system that produce a given set of observed measurements. Recent work has shown impressive results using deep learning, but we note that there is a trade-off between model performance and computational time. For some applications, the computational time at inference for the best performing inverse modeling method may be overly prohibitive to its use. We present a new method that leverages multiple manifolds as a mixture of backward (e.g., inverse) models in a forward-backward model architecture. These multiple backwards models all share a common forward model, and their training is mitigated by generating training examples from the forward model. The proposed method thus has two innovations: 1) the multiple Manifold Mixture Network (MMN) architecture, and 2) the training procedure involving augmenting backward model training data using the forward model. We demonstrate the advantages of our method by comparing to several baselines on four benchmark inverse problems, and we furthermore provide analysis to motivate its design. 
\end{abstract}

\section{Introduction}
We consider the goal of solving generic inverse problems, which arise in a multitude of scientific fields. Suppose we have an observable variable $y \in \mathbb{R}^M$, which manifests from a mapping of a forward process over a control variable $x \in \mathbb{R}^D$. We write this forward mapping as:

\begin{equation}
    y = f(x)
    \label{eq:fwd_mapping}
\end{equation}

\noindent For many scientific endeavors, while $y$ may be a measurement of a system, the variable of greater interest is the control variable $x$. This motivates the task whereby, given $f$, we seek to determine the inverse model, $g=f^{-1}(y)$, such that:

\begin{equation}
    x = f^{-1} (y)
    \label{eq:bwd_mapping}
\end{equation}

\noindent which we shall also refer to as the backward process. For many real physical systems, this inverse process is ill-posed, in that there can be several  (or sufficiently similar) settings of $x$ which correspond to a single observed $y$ (e.g., non-uniqueness of solutions).  This "one-to-many" aspect poses a particular challenge for deep learning approaches to learning inverse models, as conventional deep neural networks (DNNs) only predict a single output for a given input, and furthermore, the methods used to train DNNs (e.g., gradient descent) typically do not account for non-uniqueness. 

A variety of approaches have been proposed to address ill-posed problems of this kind \cite{aster2018parameter}, however, we focus in this work on recent DNN-based approaches, termed here as deep inverse models (DIMs) \cite{ardizzone2019analyzing, ren2020benchmarking}. Recent DIMs often fall into one of two major classes that share common advantages and disadvantages: generative and iterative. Generative models address non-uniqueness by learning a conditional probability distribution $Pr(x|y)$, and then sampling candidate solutions from this distribution. DIMS of this kind include the Variational Autoencoders (cVAEs) \cite{kingma2013auto}, Generative Adversarial Networks (GANs) \cite{salimans2016improved}, conditional Invertible Neural Networks (cINNs) \cite{kruse2019benchmarking}. The other class of DIM, iterative models, involve randomly sampling a large set of solution candidates (e.g., from a uniform distribution), and then iteratively refining these solutions (e.g., using gradient descent, or evolutionary approaches). DIMs of this kind include evolutionary algorithms \cite{zhang2020machine, johnson_genetic1997, rahmat_samii_GA_2003} and the Neural-Adjoint \cite{ren2020benchmarking}.  

Generative models are computationally fast; however, they are often less accurate than Iterative methods. As reported in \cite{ren2020benchmarking}, these models often produce solutions near - but not precisely -- the true solution. Because they are probabilistic, they are unlikely to sample exact solutions, even when the conditional distribution places substantial probability mass over the exact solution. By contrast, iterative models are highly accurate, but suffer from substantially greater computational burden, due to the need for an iterative (sequential) refinement procedure at test time.  Sequential methods are often several orders of magnitude slower than generative counterparts. For many practical problems, such as those in scientific computing (e.g., materials design \cite{khatib2021deep}, geophysics \cite{snieder1999inverse}, astronomy \cite{lucy1994optimum}, oceanography \cite{wunsch1996ocean}) this burden can be tolerated, either because real-time operation is unnecessary, or because the computational cost of the model is insignificant compared to its accuracy advantages.  For many problems though, this computational burden is unacceptable, making Generative models as the only viable solution. 

In this work, we propose Mixture Manifold Networks (MMNs), which achieve accuracy close to Iterative methods such as NA, but with computational burden similar to Generative models.  MMN is built upon an existing model, termed the Tandem \cite{liu2018training}. By contrast to Iterative and Generative models, the Tandem addresses non-uniqueness by isolating a single inverse solution for each setting of $y$, and then learning a direct mapping to that single solution, analogous to standard regression.  As a result, the Tandem model requires no iterative refinement of its solutions, but can also directly regress onto exact solutions, potentially yielding much higher accuracy than Generative models.  Indeed, recently the Tandem was found to be computationally fast, and achieve relatively high accuracy compared to other DIMs, \textit{if all models are limited to a single solution proposal} \cite{ren2020benchmarking}.  One drawback of the Tandem, however, is that it can only produce a single candidate solution, and if this solution is inaccurate, then it cannot propose additional candidates.  As a result, it tends to perform poorly compared to Iterative and Generative methods when a large number of solution proposals can be considered.  

To improve a Tandem-esque model, we seek to endow it with the capability of producing more than one candidate solution. Since the Tandem model learns a direct mapping from each $y$ to a single solution, we consider these solutions as comprising a manifold in the output space. The shape of this manifold will have been influenced by stochasticity inherent to the model training process, such as the random initialization of the model parameters and random order of training examples. Thus, we believe that if several such manifolds are learned, each will describe a different set of solutions for the same inputs $y$. While these manifolds of solutions may be similar for some $y$, it is possible -- and likely -- that certain manifolds will be better fit for different sets of $y$ than others. With this in mind, we posit that a Tandem-esque model can be made more flexible and effective by leveraging a mixture of manifolds and choosing candidate solutions from among these manifolds. We thus name our method the \textit{\underline{M}ixture \underline{M}anifold \underline{N}etwork} (MMN). 



In addition to proposing the MMN architecture, we enhance our method with a data augmentation scheme for manifold-based inverse modeling. Our second proposal comes from our observation that in the forward-backward model paradigm, the forward model is easier to learn than the backward model: the forward model does not exhibit a one-to-many challenge and generally has a more-smooth output space. This renders the forward model more capable of interpolating than the backward model. We devise a method to leverage this characteristic of the forward model to improve MMN performance.


In summary, the three primary contributions of this work are as follows:
\begin{enumerate}
    \item \textit{The Mixture Manifold Network (MMN) method for inverse modeling}. 
    MMN has performance that nears that of slower iterative models, but exhibits speed similar to generative models, thus bringing higher accuracy to more real-time applications.
    \item \textit{We demonstrate the performance of MMN on a series of benchmark inverse modeling problems.} MMN outperforms several inverse modeling baselines, and it approaches the performance of the best-performing inverse modeling method we are aware of: the Neural Adjoint (NA).  MMN is drastically faster at inference than NA. 
    \item \textit{We provide analysis that lends insight into why MMN is an effective inverse modeling method.}
\end{enumerate}
\section{Related Work}
\label{sec:related_work}
Contemporary DIM methods often fall into either a \textit{generative} or \textit{iterative} paradigm, with the Tandem method falling into its own separate paradigm. We discuss the Tandem more fully in Section \ref{sec:methods}, but mention here that in this approach, a neural forward model is first trained and then its parameters are fixed before the forward model is connected to the backward model, which is then trained in end-to-end fashion. Connecting the forward and backward models thereby regulates the noisy gradient caused by one-to-manyness. 


Another class of end-to-end approaches is to model the conditional posterior distribution of the control variables given the observable variables, $p(x|y)$.
These approaches leverage variational methods, and include models such as Variational Auto-Encoders (VAEs), Mixture Density Networks (MDNs) \cite{bishop1994mixture}, Invertible Neural Networks (INNs), and conditional INNs (cINNs).

VAEs use an encoder-decoder structure, with the encoder modeling $x$ into a Gaussian-distributed random variable $z$ conditioned on $y$, and the decoder going back from the latent variable $z$ to $\hat{x}$. The VAE is trained by minimizing the evidence-lower-bound (ELBO) as well as a L2 MSE reconstruction loss between original input $x$ and $\hat{x}$. 
MDNs assume a Gaussian mixture for the posterior $p(x | y)$, with the parameters of Gaussians (as well as the mixing proportions) predicted by a feedforward neural network, and the number of components in the mixture being selected as a hyperparameter. At inference for a particular $y$, $\hat{x}$ is sampled from the learned mixture distribution. 
Like, VAEs, INNs also assume latent Gaussian variables $z$, but now \textit{bijective} mappings are learned between $x$ and pairs $(y, z)$. INNs can be trained by minimizing a supervised reconstruction loss and Maximum Mean Discrepancy (MMD) loss, or alternatively, a maximum likelihood estimate (MLE) loss used to enforce Gaussianity for the distribution of $z$. For proposing inverse solutions $y$, values of $z$ are sampled from a Gaussian distribution, and the pairs $(y, z)$ are mapped by the network to an estimate $x$. Finally, cINNs modify the INN structure in that they learn bijective mappings between $x$ and $z$ conditioned on $y$, rather than between $x$ and the pair $(y, z)$.  

The last class of DIM approaches we consider here are iterative methods, such as the Genetic Algorithm (GA) and Neural-Adjoint (NA). Under this paradigm, the DIM will make an initial set of solution proposals, and then from these iteratively search for better solutions. The NA uses a neural approximation of the forward process $\hat{f}$, and searches for optimal solutions $\hat{x}$ using the gradient $\partial \hat{f} / \partial x$. The GA, on the other hand, is a gradient-free method, and instead searches for solutions by selecting and mutating populations of candidate solutions using evolutionary fitness rules.

Many of the methods above have recently been benchmarked for different inverse problems. Kruse \cite{kruse2019benchmarking} benchmarked the stochastic (e.g., posterior modeling) approachs on two simple kinematic inverse problems, and Ren \cite{ren2020benchmarking} further benchmarked those methods on two additional problems. That latter study also benchmarked the Tandem \cite{liu2018training} model and their proposed NA on the same four inverse modeling tasks. 


\section{Methods}
\label{sec:methods}
\begin{figure}[h]
\centering
\includegraphics[width=0.425\textwidth]{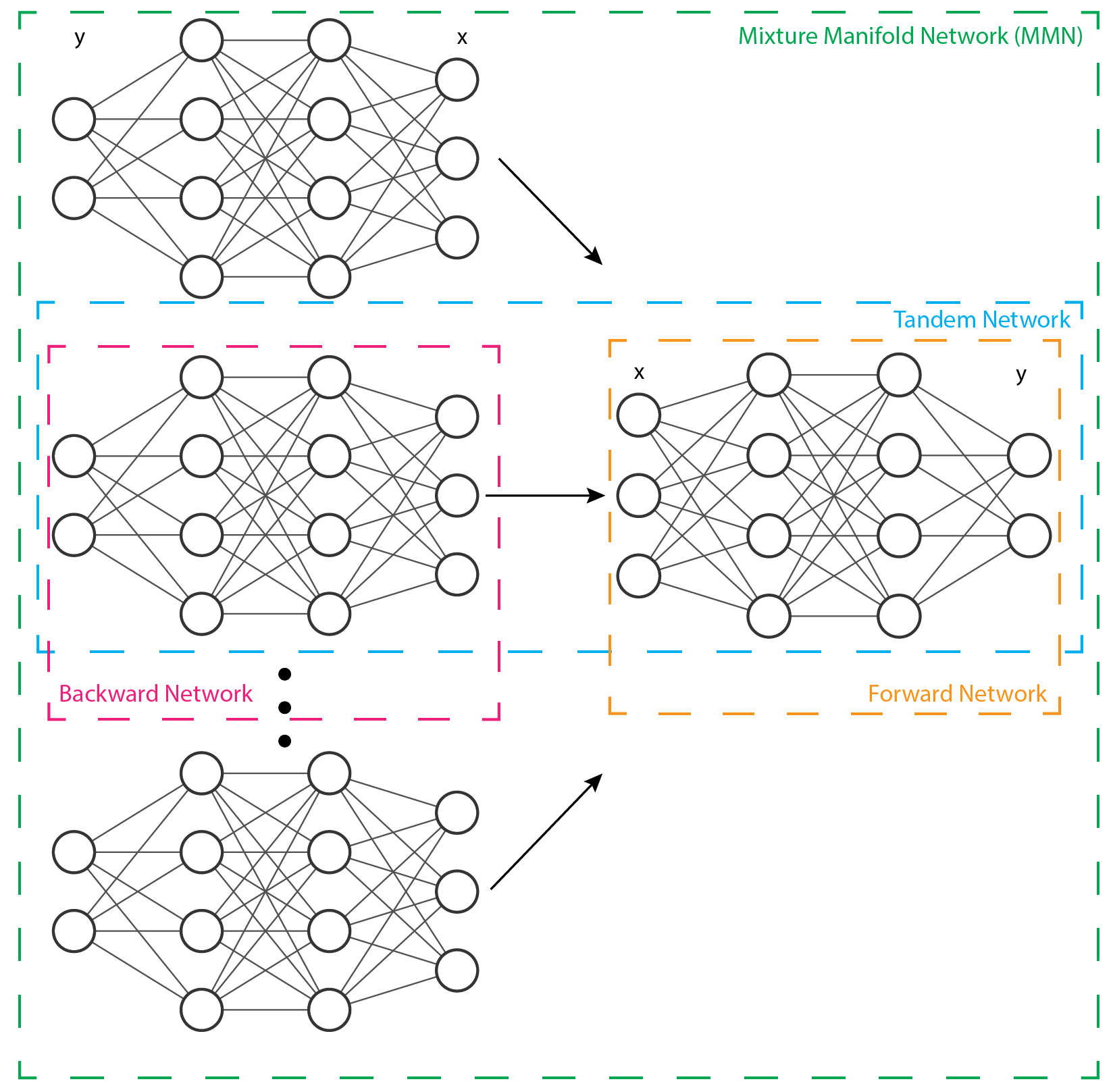} 
\caption{Model architecture illustration. The yellow box represents the forward network that maps from control variable $x$ to observation $y$. The pink box represents the backward network that maps the inverse relationship from $y$ to $x$. Connecting them forms the Tandem network, depicted in the blue box. Our proposed MMN can be seen as an extension of Tandem network, where multiple backward models are trained corresponding to single trained forward model, and this is illustrated in the green box.}
\label{fig:network_architecture}
\end{figure}

The MMN method bears resemblance to the Tandem, particularly in the case that a single manifold is used (effectively reduces to a Tandem network). We thus first review the Tandem's architecture and training procedure before describing our proposed MMN, and both are illustrated in Figure \ref{fig:network_architecture}. The methods require a dataset upon which to train, consisting of sample pairs $\{x_n, y_n\}_{n=1}^N$ from the true process $f$. 

\subsection{Tandem Model}
\label{sec:tandem}
The Tandem architecture consists of two neural networks. One network approximates the forward process, while the other models the backward process. 
Supposing that the estimate, $\hat{f}$, of the forward process is parameterized by $\theta_f$, then for each $x_n$ in the training data, the model predicts a corresponding output $\hat{y_n}$:

\begin{equation}
    \hat{y}_n = \hat{f} (x_n ; \theta_{f})
    \label{eq:fwd_model}
\end{equation}

\noindent The parameters $\theta_f$ of the forward model are learned by optimizing a standard regression loss; here, we use the squared error loss between the true $y_n$ and the estimate $\hat{y}_n$:

\begin{equation}
    \mathcal{L}_{fwd} = \sum_{n=1}^N \Big| \Big| y_n - \hat{y}_n \Big| \Big|_2^2
    \label{eq:fwd_model_loss}
\end{equation}

After learning the forward model $\hat{f}$, its parameters are fixed (e.g., ``frozen") before estimating the inverse model $\hat{f}^{-1}$, the second network composing the Tandem. For notational convenience, we will denote $\hat{g} = \hat{f}^{-1}$:

\begin{equation}
    \hat{x}_n = \hat{f}^{-1} (y_n) = \hat{g} (y_n ; \theta_g)
    \label{eq:bwd_model}
\end{equation}

\noindent where $\theta_g$ parameterizes the network $\hat{g}$.  These parameters are learned by minimizing the following loss:

\begin{equation}
    \mathcal{L}_{g} = \sum_{n=1}^N \Big| \Big|  \hat{f} (\hat{g}(y_n)) - y_n \Big| \Big|^2_2
    \label{eq:bwd_model_loss}
\end{equation}

\noindent where we reiterate that the parameters of $\hat{f}$ have been frozen. 

In addition to the loss described in \ref{eq:bwd_model_loss}, we employ a boundary loss proposed by \cite{ren2020benchmarking}. This loss was introduced to mitigate a challenge with their Neural Adjoint in that it would frequently converge to solutions outside the domain of the training data. They posit the reason for this is that outside of the training data domain (for points $x$), the estimated forward model $\hat{f}$ becomes highly inaccurate and then tends to erroneously predict that $x$-values outside of the space of $\mathcal{X}$ are near the desired solutions for $y$. To address this undesirable behavior, an additional loss term is included to encourage the models to identify solutions within the bounds of the training data (e.g., within the space $\mathcal{X}$). The formulation of this loss term, $\mathcal{L}_{bnd}$ is given by:

\begin{equation} \label{eq:boundary_loss}
    \mathcal{L}_{bnd}(x) = 
    \begin{cases}
        x - x_{max}, & \text{if } x \geq x_{max}\\
        0, & \text{if } x_{min} \leq x \leq x_{max}\\
        x_{min} - x, & \text{if } x \leq x_{min}
    \end{cases}
\end{equation}

\noindent which is eqeivalent to $\mathcal{L}_{bnd} = \mathrm{ReLU} \big( |x - \mu_x | - \frac{1}{2} R_x \big)$ for uniform distributions where $\mu_x$ and $R_x$ are the mean and range of the training data, respectively, and $\mathrm{ReLU}$ is the rectified-linear-unit commonly used as an activation function in neural networks.  For data distributions without hard bounds on $x$, the range was defined to the be the interval of $95\%$ probability.  Though the boundary loss was proposed for training the Neural Adjoint inverse models, the authors of \cite{ren2020benchmarking} demonstrate that including $\mathcal{L}_{bnd}$ in the training of Tandem backward models also improves their performance. Thus, when training the backward component of a Tandem model, we use a combination loss from Equations \ref{eq:bwd_model_loss} and \ref{eq:boundary_loss}:

\begin{equation}
    \mathcal{L}_{bwd} = \mathcal{L}_{g} + \gamma \mathcal{L}_{bnd}
    \label{eq:total_bwd_model_loss}
\end{equation}

\noindent where the hyperparameter $\gamma$ is used to balance the relative importance of the two losses in the combination. 

\subsection{Mixture Manifold Network}
We propose to extend the flexibility of a forward-backward model architecture like the Tandem by leveraging multiple backwards models associated with a single forward model. Since the outputs from each of these backward models can be considered as lying on a separate \textit{manifold}, and since we propose to use a \textit{mixture} of such manifolds, we have named our method the \textit{Mixture Manifold Network}.

As in the Tandem procedure previously described, the MMN method begins by estimating a single forward model $\hat{f}$ as an approximation of the forward process described in Equation \ref{eq:fwd_mapping}. This estimate is learned by optimizing the loss in Equation \ref{eq:fwd_model_loss} with predictions as in Equation \ref{eq:fwd_model}.  

Also as in the Tandem method, after learning the forward model $\hat{f}$, its parameters are frozen before learning subsequent pieces of the network. Whereas the Tandem model learns a single inverse model $\hat{g} = \hat{f}^{-1}$, our Mixture Manifold Network learns a set of $K$ inverse models, $\{\hat{g}^k\}_{k=1}^K$. Crucially, each backward model $\hat{g}^k$ is learned separately as an approximation for $f^{-1}$ (Equation \ref{eq:bwd_mapping}), using the same training data and using the combination of backward model loss and boundary loss in Equation \ref{eq:total_bwd_model_loss}. Each backward model's predictions are can be described by:

\begin{equation}
    \hat{x}_n^k = \hat{g}^k (y_n ; \theta_g^k)
    \label{eq:bwd_model_MMN}
\end{equation}

\begin{figure}[h]
\centering
\includegraphics[width=0.425\textwidth]{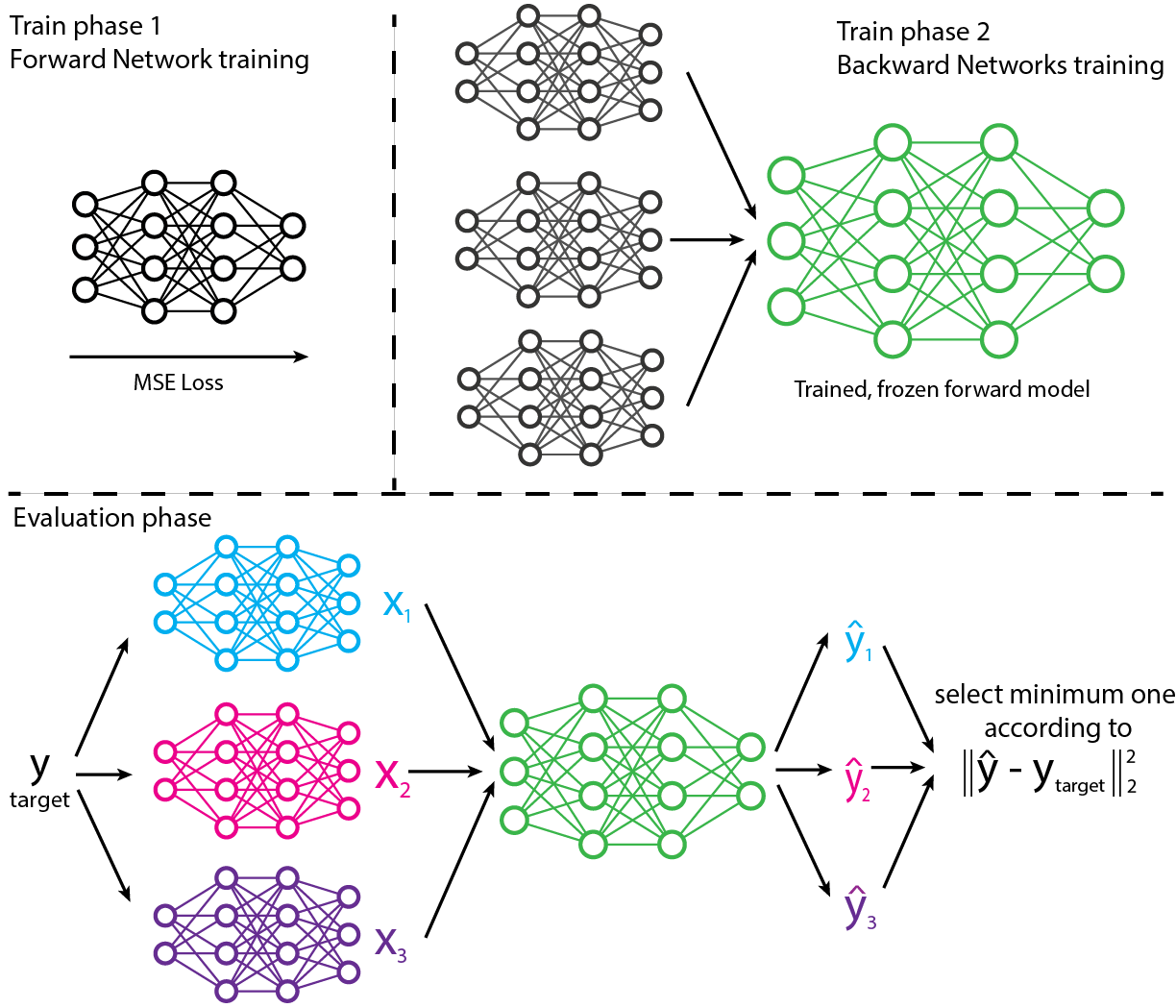} 
\caption{Illustration the training and testing phases of MMN. The forward and backward networks are trained sequentially, with the trained forward network connected at the end of the backward network during the backward model's training. During evaluation, the backward networks pass their outputs through the forward model and the best design is chosen according to Equation \ref{eq:MMN_inference}}
\label{fig:train_eval_phases}
\end{figure}

To leverage a mixture of $K$ trained backward models, we devise a new inference scheme that is critical to the effectiveness of MMN. The scheme is a procedure whereby for each $y_n$ MMN essentially ``chooses" which of the manifolds in its mixture it will use to propose a solution $\hat{x}_n$; it is illustrated in Figure \ref{fig:train_eval_phases}. Consider that each backwards model in MMN has predicted a $\hat{x}_n^k$ using Equation \ref{eq:bwd_model_MMN}, yielding a set $\{\hat{x}_n^k\}_{k=1}^K$. Since the same forward model $\hat{f}$ is tied to each backward model, the forward model can be used to assess which of the proposed solutions is the best.  This can be accomplished by using the forward model to map each $\hat{x}_n^k$ back into $\mathcal{Y}$-space and comparing to the original test point $y_n$.  Whichever proposed solution yields an output closest to $y_n$ is selected, which we formalize as:

\begin{equation}
    \hat{x}_n = \mathrm{argmin}_{\hat{x}_n^k} \Big| \Big| \hat{f}(\hat{x}_n^k) - y_n \Big| \Big|^2_2
    \label{eq:MMN_inference}
\end{equation}

\subsection{Sampling and Forward Model Generation}
\label{sec:fwd_model_sampling}

Whereas we have thus far described the MMN \textit{architecture} and an inference procedure to effectively leverage that architecture, we describe here a proposed innovation to mitigate the training of efficacious backward models $\hat{g}^k$.  Our proposal is to \textit{augment} the data used to train the backward models comprising a MMN by leveraging the effectiveness of the forward model $\hat{f}$. This augmentation involves sampling points from the input space $\mathcal{X}$ and then to use the learned forward model $\hat{f}$ to map those points into the output space $\mathcal{Y}$ (e.g., Equation \ref{eq:fwd_model}). These pseudo-points which have been ``generated" by the forward model are then used as the training data for the backward models in the MMN. More formally, we propose to assemble a new set of training data $\{\tilde{x}_n, \tilde{y}_n\}_{n=1}^{N'}$ with which to train the backward models $\hat{g}^k$. This new set of training data is generated according to the following:


\begin{equation}
    \tilde{y} = \hat{f} (\tilde{x}); \ \tilde{x} \sim \hat{p}(x)
    \label{eq:fwd_model_sampling}
\end{equation}

\noindent where $\hat{p}(x)$ is a placeholder to indicate that the points $\tilde{x}$ are sampled from a distribution that bears \textit{resemblance} to the distribution describing the true data-generating process for $x$, but is perhaps not the same distribution. The most crucial aspect about the sampling distribution $\hat{p}$ is that it shares the same support (e.g., bounds on $x$) as the true distribution $p$.  In some cases, we may know $p(x)$ exactly -- as in the Sine Wave and Robotic Arm benchmarks -- and so we can sample directly from that distribution.  In other cases, like the Meta-Material and Shell problems, generic distributions (e.g., uniform, Gaussian) with judiciously chosen bounds and means can be chosen. 

\subsection{Evaluation -- Re-simulation Error}
\label{sec:evaluation}

We measure model performance using ``re-simulation" error, $r = \mathcal{L} \big(f(\hat{x}), y\big)$, which seeks to capture the discrepancy between the true solution $y$ and the ``re-simulated" value of $y$ obtained by passing the estimate $\hat{x}$ through the true forward model (also known as the simulator). We follow the convention of using mean-squared error (MSE) to measure this discrepancy. \cite{ren2020benchmarking} extended resimulation error to account for proposal of $T$ solutions from a model. Here, $T$ is the number of times the simulator $f$ is used, which may be computationally expensive (e.g., a material design physics-based simulation). 




\section{Benchmark Problems}
We consider four benchmark tasks to demonstrate our MMN inverse modeling method; these are summarized in Table \ref{table:benchmarks}. 
We include a popular task -- the Robotic Arm -- which was introduced in \cite{kruse2019benchmarking} and further benchmarked in \cite{ren2020benchmarking}. That latter study also benchmarked the Sine Wave (which it introduced) and the Meta-Material \cite{nadell2019deep, deng2021neural} tasks. For our fourth task, we have adopted the Shell \cite{peurifoy2018nanophotonic}, which was benchmarked in \cite{ren2022inverse}. 

We have selected these benchmarks for several reasons. First, while each demonstrates non-uniqueness of solutions, they exhibit a range of technical difficulty, with the Sine Wave being the easiest due to its low dimensional $x$ and $y$, and the Meta-Material and Shell problems being harder due to their relatively higher dimensionality. Second, they encompass a breadth of application, with the Sine Wave being somewhat of a toy problem useful for close analysis, Robotic Arm being a kinematic problem with possible application for control systems, and both Meta-Material and Shell being from two different sub-fields of artificial electromagnetic material design. We believe that choosing this range of problems helps ensure that our findings for MMN generalize to other possible inverse problems. Last, each of these tasks has been benchmarked in a previous study, which helps us maintain continuity with the community. 

For each inverse task, we use the same experimental designs as from previous benchmark studies, including the simulator (e.g., forward model) parameters, simulator sampling procedures, and training/testing splits. Details not included in Table \ref{table:benchmarks} can be found in the studies cited above and in our supplement. 

\begin{table}
	\small
	\centering
	\begin{tabular}{c c c c c c}
    	\toprule
		 & Sine & Arm & Meta-Mat & Shell & \\
        \midrule
		Dim(x) & 2 & 4 & 8 & 8 \\
		Dim(y) & 1 & 2 & 300 & 201 \\ 
		Num. Train & 8000 & 8000 & 8000 & 40,000 \\ 
		Num. Val. & 2000 & 2000 & 2000 & 10,000 \\
		Num. Test & 1000 & 1000 & 1000 & 500 \\
	    \bottomrule 
	\end{tabular}
	\caption{Summary of our inverse problem datasets. Note we cover both cases that $Dim_x > Dim_y$ and $Dim_y > Dim_x$.}
	\label{table:benchmarks}
\end{table}

\section{Experimental Design and Results}
\label{sec:experiments}

We show experiments and results on the above benchmark problems to demonstrate the performance of our proposed MMN method and compare it to an array of previous inverse modeling methods. For the baselines, we find mostly consistent results with previous benchmarking studies, and we find that our MMN outperforms generative model baselines without the severe computational time of iterative models. Thus, we argue that our MMN method provides an alternative that balances performance with inference speed. 


We strive to follow the experimental design of \cite{ren2020benchmarking} to the closest extent possible, as they have recently benchmarked baseline deep inverse models on some of the same problems (note that their design, in turn, follows that of \cite{kruse2019benchmarking}). For the Shell problem, we strive to follow \cite{ren2022inverse}, as that problem has recently been benchmarked therein. For each benchmark problem, each model used the same training and testing data, batch size, and criteria to stop training. Model architectures and hyperparameters we selected to align with previous studies, and full implementation details can be found in the supplementary material. 

For MMN experimentation, we use the the same forward and backward model architectures as used for the Tandem model for each benchmark problem. We use $K=6$ manifolds (backward models) for each MMN. 
Each of the $K$ backward models were trained using \textit{only} augmented data generated using the forward model sampling procedure described in the Methods section. The distributions used for this sampling differed between benchmarks, and they are outlined in the supplement. For the Sine Wave, Robotic Arm, and Meta-Material benchmarks, we sampled 40,000 points with which to train, and for the Shell benchmark, we sampled 250,000 (in the next section, we show performance as the number of generated points varies). 


\begin{table}
	\small
	\centering
	\begin{tabular}{c c  c  c  c }
        \toprule
		& Sine & Arm & Meta-Mat. & Shell \\
		\midrule
		MDN & 3.68e-1 & 4.41e-3 & 2.65e-3 & 1.07e-1 \\
		INN & 6.78e-1 & 6.26e-3 & 2.84e-2  & 1.79e-2 \\ 
		cINN & 3.43e-1 & 4.17e-3 & 4.68e-3 & 7.33e-2 \\ 
		cVAE & 8.18e-1 & 2.32e-2 & 1.20e-2 &  7.51e-3 \\
		TD &  1.27e-2 & 2.38e-3 & 1.36e-3 & 7.41e-3 \\ 
		GA & 8.15e-3 & 1.65e-4 & 2.66e-4 & 2.89e-3 \\
		NA & 3.25e-3 & \textbf{1.17e-4} & \textbf{1.97e-4} & \textbf{2.88e-3} \\ 
		MMN &\textbf{1.32e-3} & 4.33e-4 & 3.50e-4 & 3.75e-3 \\
	    \bottomrule	    
	\end{tabular}
	\caption{Average T=1 Re-Simulation Error}
	\label{tab:T_1_resim_error}
\end{table}

\begin{figure*}[t]
\centering
\includegraphics[width=0.73\textwidth]{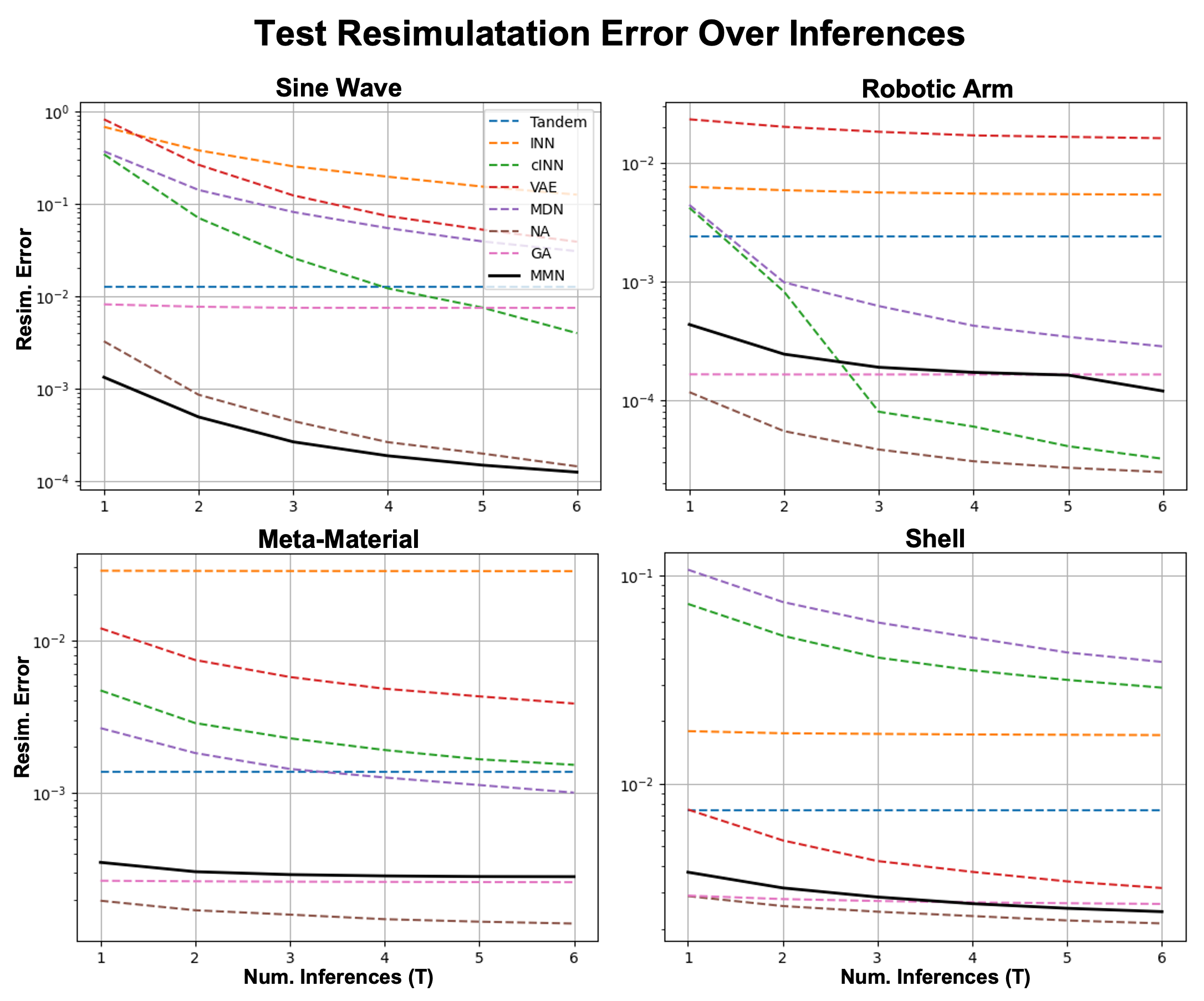} 
\caption{Re-Simulation error over Number of Inferences. We highlight our MMN method with the solid, black line, while baseline methods are shown with dashed lines.}
\label{fig:resim_vs_T}
\end{figure*}


Our main experimental results are presented in Table \ref{tab:T_1_resim_error}, which shows the re-simulation error for MMN and all baseline methods for the case that there are $T=1$ solution proposals by each model. Here, we find that MMN is either the best performing method (in the case of the Sine Wave problem) or is the second-best performing method to NA, which performs so well at an expense of heavy computation during evaluation. We thus find that on all four benchmarking problems, MMN outperforms the Tandem model -- to which it bears the most resemblance -- as well as all of the probabilistic models (MDN, INN, cINN, and cVAE).

In addition to evaluation for $T=1$ solution proposals from each model, we can measure re-simulation error performance as a function of $T$, as in \cite{ren2020benchmarking}. A limiting aspect of Tandem performance is its incapability of predicting multiple solutions, but our MMN method -- while a deterministic model -- does admit multiple solutions. To be exact, MMN can propose up to $T=K$ different solutions, as we take the solution from each manifold as a separate prediction. In Figure \ref{fig:resim_vs_T} we show re-simulation error as a function of $T$ for each model on each benchmark problem.  
We also provide in Table $\ref{tab:T_6_resim_error}$ the re-simulation error for $T=6$ inferences. With multiple proposed solutions, our MMN still outperforms all but the iterative method baselines, to which MMN still remains competitive. 


\begin{table}
	\small
	\centering
	\begin{tabular}{c c  c  c  c }
        \toprule
		& Sine & Arm & Meta-Mat. & Shell \\
		\midrule
		MDN & 3.06e-2 & 2.83e-4 &  1.00e-3 & 3.85e-2 \\
		INN & 1.25e-1 & 5.38e-3 & 2.83e-3 & 1.71e-2 \\ 		
		cINN & 3.97e-3 & 3.20e-5 & 1.52e-3 & 2.90e-2 \\ 
		cVAE & 3.88e-2 & 2.61e-2 & 3.84e-3 & 3.15e-3 \\
		TD & 1.27e-2 & 2.38e-3 & 1.36e-3 & 7.41e-3 \\ 
		GA & 7.47e-3 & 1.64e-4 & 2.60e-4 & 2.64e-3 \\
		NA & 1.44e-4 & \textbf{2.47e-5} & \textbf{1.39e-4} & \textbf{2.13e-3} \\ 
		MMN & \textbf{1.25e-4} & 1.19e-4 & 2.82e-4 & 2.42e-3 \\
	    \bottomrule	    
	\end{tabular}
	\caption{Average T=6 Re-Simulation Error}
	\label{tab:T_6_resim_error}
\end{table}

We have argued that a possible limitation of the iterative NA and GA methods which outperform our MMN is that they require severe inference time, which may be prohibitive for real-time applications or other situations when fast inference is required. The MMN method does not exhibit such inference times, which we show in Table \ref{tab:inference_time}, presenting the total inference time required in seconds for each model to propose 1000 solutions to each benchmark problem. Clearly, both the NA and GA methods require significantly more inference time than any of the other baseline methods, and this is a direct consequence of the iterative nature of their solution finding.  This inference time drawback is also compounded when the dimensionality of the inverse problem is larger, as can be seen for specifically the Meta-Material and Shell benchmarks.


\begin{table}
	\small
	\centering
	\begin{tabular}{c c  c  c  c }
        \toprule
		& Sine & Arm & Meta-Mat. & Shell \\
		\midrule
		MDN & 0.27 & 0.27 & 0.49 & 0.18 \\
		INN & 0.03 & 0.03 & 0.25 & 0.20 \\ 		
		cINN & 0.03 & 0.03 & 0.26 & 0.43 \\ 
		cVAE & 0.02 & 0.02 & 0.24 & 0.18 \\
		TD & 0.02 & 0.02 & 0.23 & 0.20 \\ 
		GA & 28.28 & 28.20 & 97.98 & 71.93 \\
		NA & 1.83 & 1.84 & 80.44 & 54.57\\ 
		MMN & 0.13 & 0.15 & 3.26 & 1.22\\ 
	    \bottomrule	    
	\end{tabular}
	\caption{Inference Time in seconds for 1,000 solutions}
	\label{tab:inference_time}
\end{table}


\section{MMN Analysis}
\label{sec:mmn_analysis}

In this section, we further characterize how our proposed MMN method performs. In particular, we examine how both of the innovations in MMN change its performance in terms of resimulation error. Our results demonstrate that each innovation individually -- using multiple manifolds and training using forward-model generated points -- improves performance, suggesting that both are individually beneficial.  

\subsection{On Use of Multiple Manifolds}

\begin{figure}
\centering
\includegraphics[width=0.425\textwidth]{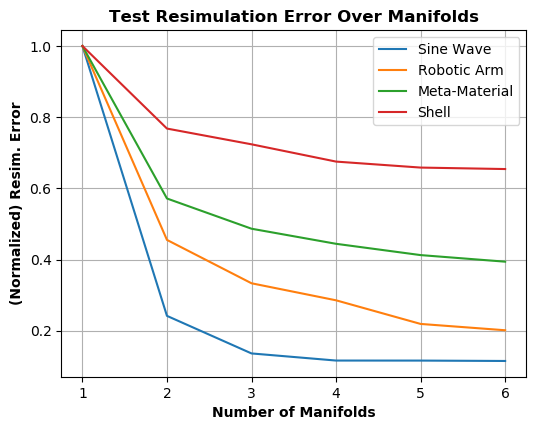}
\caption{Resimulation error over number of manifolds for MMN on the test set for each benchmark problem. Note that resimulation error here has been normalized by the resimulation error for a single-manifold model, so as to put each benchmark's results over number of manifolds on the same scale and facilitate viewing.}
\label{fig:resim_over_num_manifolds}
\end{figure}

We provide evidence that our proposed architecture for the Mixture Manifold Network -- separate from using forward model point generation -- is beneficial to modeling performance. 
To demonstrate that using multiple backwards models offers flexibility, we show the performance, in terms of $T=1$ resimulation error, improves as successive manifolds are included in MMN. We present this for all benchmark problems in Figure \ref{fig:resim_over_num_manifolds}, where the resimulation error across number of manifolds has been normalized by single-manifold performance so that results for all benchmarks can be more-easily viewed on the same axes. 

For all benchmarks, we observe a decrease in resimulation error when $K>1$, and furthermore, this decrease is monotonic with increasing $K$. This indicates that as successive manifolds are added to MMN, the model's overall performance increases.  We believe that this experimentally demonstrates evidence for our intuition that using multiple manifolds allows MMN to ``cover" different regions of $\mathcal{X}$-space with different manifolds, and that a well-trained forward model can discern which manifold is the most appropriate to use depending on the input point. Thus, as more manifolds are added to MMN, the forward model has more flexibility in its selection of the best manifold. 

\subsection{On Use of Sample Generation}

\begin{figure}[t]
\centering
\includegraphics[width=0.425\textwidth]{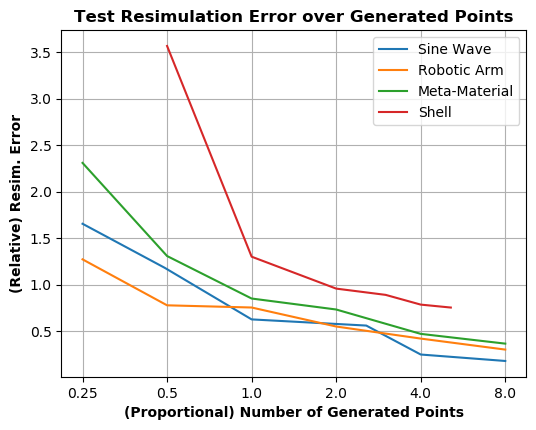}
\caption{Resimulation error on test over number of points sampled and generated by the MMN forward model for each benchmark problem.
Note that both axes have been adjusted to facilitate assessing results across all benchmarks. The x-axis, number of points sampled/generated, is in terms of proportionality to the number of real points used to train baseline models (and used to train the forward model itself). The resimulation error, shown on the y-axis, is relative to the error obtained by a (single-manifold) MMN with backward models trained on real data.}
\label{fig:resim_over_num_points_sampled_1}
\end{figure}

We now demonstrate that training backward models using points that have been sampled and pushed through the forward model is also beneficial for MMN performance. Again, we consider performance in terms of $T=1$ resimulation error, and now we show that as the number of points generated by the forward model increases, so too does the performance of MMN. This result is presented, for every benchmark problem, in Figure \ref{fig:resim_over_num_points_sampled_1}. To present all benchmarks in the same space, we have again adjusted the axes of this results figure. The x-axis shows the number of generated points proportional to the number of real points used to train the baseline models, and the y-axis is relative to the resimulation error obtained by MMN with the (single) backward model trained on real data. 

In addition to the resimulation error decreasing as the number of generated points increases, it is almost more important to note that for each benchmark, there is a certain number of generated points at which MMN outperforms a baseline with backward models trained on real data (e.g., the curves for all benchmarks fall below $\bar{r} = 1$).

\section{Conclusions}
\label{sec:conclusions}

In this work, we have proposed a new method for approximating solutions to inverse problems using deep neural networks, and we have assessed the performance of our method against seven modern deep inverse models on four benchmark tasks.  Our method, the Mixture Manifold Network, proposes a solution from among multiple backward models that have all been trained to correspond to the same forward model, and it can furthermore at training leverage augmented data points generated from the forward model. We find that our method performs comparably to iterative methods and outperforms all other baselines. MMN, however, exhibits dramatically faster inference time than the iterative methods do.
Thus, for applications where inference time may be a limiting concern (e.g., real-time control systems or rapid material design), we believe our MMN offers a promising approach to addressing such inverse problems. 

\bibliography{mmn_bib}
\newpage
\appendix

\section{Appendices}
Here, we provide details that were omitted from the main paper, as they are not crucial to understanding the method or our results, but they do aid in any replication efforts. These details include more information about the four benchmark problems that we use to demonstrate the efficacy of our method and compare it to several baselines, architecture and hyperparameter specifications for models we use, and details surrounding the distributions from which we sampled to use our proposed method for generating points from trained forward models.  

\section{Benchmark Problems}
We consider four benchmark tasks to demonstrate our MMN inverse modeling method; these are summarized in Table \ref{table:benchmarks}. 
We include a popular task -- the Robotic Arm -- which was introduced in \cite{kruse2019benchmarking} and further benchmarked in \cite{ren2020benchmarking}. That latter study also benchmarked the Sine Wave (which it introduced) and the Meta-Material (\cite{nadell2019deep} tasks. For our fourth task, we have adopted the Shell \cite{peurifoy2018nanophotonic}, which was benchmarked in \cite{ren2022inverse}.  For more details beyond those we provide below, we encourage interested readers to see the benchmark studies above. 

\subsection{Sine Wave}
The sine wave benchmark problem consists of a sinusoidal function in two dimensions described by:
\begin{equation*}
    y = \mathrm{sin}(3\pi x_1) + \mathrm{cos}(3\pi x_2)
\end{equation*}
\noindent To generate data for this problem, points are independently sampled for each dimension of $x$ from a uniform distribution on the interval $[-1, 1]$. As noted in \cite{ren2020benchmarking}, despite this problem's simple formulation, it exhibits the non-uniqueness in the backward process that renders it a challenge for most deep inverse models.  Furthermore, the 2-dimensional input space is conducive to visualizing solutions proposed by each inverse model, making this a useful benchmark for the analysis of their errors. 

\subsection{Robotic Arm}
Also called simply ``Inverse Kinematics", this benchmark task was introduced in \cite{ardizzone2019analyzing}, and it simulates the articulation of a multi-jointed robotic arm as it moves in 2D space. The input $x \in \mathbb{R}^4$ of the forward process consists of the starting height $x_1$ of the arm and the angles -- $x_2, x_3, x_4$ -- of the arm's three joints.  The output of the forward process $y \in \mathbb{R}^2$ is the 2-D position of the arm's end-point. The two coordinates of $y$ are given by:

{\small 
\begin{align*}
    y_1 & = l_1 \mathrm{sin} (x_2) + l_2 \mathrm{sin}(x_2+x_3) + l_3 \mathrm{sin}(x_2+x_3+x_4)+x_1\\
    y_2 & = l_1 \mathrm{cos}(x_2) + l_2 \mathrm{cos}(x_2+x_3) + l_3 \mathrm{cos}(x_2 + x_3 + x_4)
\end{align*}}

\noindent with arm-segment lengths $l_1=\frac{1}{2}$, $l_2 = \frac{1}{2}$, and $l_3 = 1$.  The input $\mathbf{x}$ is drawn according to a Gaussian: $\mathbf{x} \sim \mathcal{N} (0, \boldsymbol{\sigma}^2 \cdot \mathbf{I})$, with $\boldsymbol{\sigma}^2 = [\frac{1}{16}, \frac{1}{4}, \frac{1}{4}, \frac{1}{4}]$. 

\subsection{Meta-Material}
\label{sec:meta_material}

The Meta-Material problem was introduced in \cite{nadell2019deep} and recently used for benchmarking in \cite{ren2020benchmarking}. A material design problem, the goal is to choose the geometries -- radii and heights of four cylinders -- of a meta-material to yield a desired electromagnetic reflection spectrum. The underlying mechanics of the forward process, $f(x)$, necessitate significant time and expertise to evaluate solutions using slow, iterative electromagnetic simulators. To overcome this impediment to use as a benchmarking inverse problem, \cite{ren2020benchmarking} introduced an ensemble of well-trained deep neural networks to approximate the simulator. To obtain these neural networks, they generated a large number of samples (approximately 40,000) from the (slow) electromagnetic simulator with which to train the neural simulators. Their benchmarking experiments on this problem used data from such a proxy simulator, and they provide the ensemble of networks on the Github repository associated with their work.  To perform our experiments on the Meta-Material problem, we have employed the neural simulators provided by \cite{ren2020benchmarking}. 

\begin{table}[t]
	\small
	\centering
	\begin{tabular}{c c c c c c}
    	\toprule
		 & Sine & Arm & Meta-Mat & Shell & \\
        \midrule
		Dim(x) & 2 & 4 & 8 & 8 \\
		Dim(y) & 1 & 2 & 300 & 201 \\ 
		Num. Train & 8000 & 8000 & 8000 & 40,000 \\ 
		Num. Val. & 2000 & 2000 & 2000 & 10,000 \\
		Num. Test & 1000 & 1000 & 1000 & 500 \\
	    \bottomrule 
	\end{tabular}
	\caption{Summary of our inverse problem datasets. Note we cover both cases that $Dim_x > Dim_y$ and $Dim_y > Dim_x$.}
	\label{table:benchmarks}
\end{table}

\subsection{Shell}
\label{sec:shell}
The ``Shell" problem was introduced in \cite{peurifoy2018nanophotonic}, and is another material design challenge. The goal of this problem is to again choose the geometry of a material to obtain a certain spectral response; in this case, the material is a multi-layer dialectric spherical nanoparticle comprising alternating $TiO_2$ and silica shells with tunable thickness. The original study, \cite{peurifoy2018nanophotonic}, provides an analytical simulator implemented in Matlab, and the benchmarking study that we follow, \cite{ren2022inverse}, provides a Python implementation of such a simulator. We use the latter simulator to generate the data used in our experiments and as the simulator for evaluating our resultant models. 

\section{Experimental Design: Additional Details}
\label{sec:experiments}

For the experiments presented in our main paper, we have striven to to follow as closely as possible the design provided by previous benchmarking studies. For the Sine Wave, Robotic Arm, and Meta-Material benchmarks, we follow \cite{ren2020benchmarking}, who provide code to train and evaluate baseline models, including the Neural Adjoint model. They, in turn, base their experimental design on that of \cite{kruse2019benchmarking}, at least for the Robotic Arm benchmark problem. For the Shell problem, we strive to follow \cite{ren2022inverse}, who also provided code to train and evaluate baseline models on that benchmark problem. For all benchmark problems and baselines, we have not modified the architectures or hyperparameters from what we found in the publicly available implementations, so we refer interested researchers to the studies above.

The number of training examples used for each benchmark are provided in the main paper, as well as in Table \ref{table:benchmarks} of this supplement. We additionally provide training settings common to all models and benchmarks in Table \ref{table:train_settings}

\begin{table}
	\small
	\centering
	\begin{tabular}{c c}
    	\toprule
		(Hyper)Parameter & Value \\
        \midrule
		Number Training Epochs & 500 \\
		Batch Size & 1024 \\ 
		Optimizer & Adam \\ 
		Learning Rate & 0.001 \\
		Learning Rate Schedule & Decay 1/2 at Plateau  \\
		GPU & NVIDIA Titan Xp\\
	    \bottomrule
	\end{tabular}
	\caption{Training settings common to all baseline models on all benchmarks}
	\label{table:train_settings}
\end{table}

We base the MMN architectures used for our experiments on the architectures of the Tandem \cite{liu2018training} model that were used in the benchmarking studies by \cite{ren2020benchmarking} (for Sine Wave, Robotic Arm, and Meta-Material) and \cite{ren2022inverse} (for Shell). We provide those architectures below. For all benchmarks, our final MMN results (e.g., those presented in the Results section of our main paper) were obtained using $K=6$ manifolds, which is to say $6$ backward models associated with a single forward model. For all MMN (and Tandem) models, batch normalization is used as a regularization strategy, and the rectified-linear-unit (ReLU) is used as the non-linearity between layers. 

\subsection{Sine Wave}
The forward model architecture consists of a fully-connected network with input dimensionality of $2$ and output dimensionality $1$. The number of neurons for the hidden layers are: 500, 500, 500, 500.

Each backward model for MMN also consisted of a fully-connected network, now with input dimensionality $1$ and output $2$. The number of neurons for the hidden layers was the same as above for the forward model. 

\subsection{Robotic Arm}
The forward model architecture consists of a fully-connected network with input dimensionality of $4$ and output dimensionality $2$. The number of neurons for the hidden layers are: 500, 500, 500.

Each backward model for MMN also consisted of a fully-connected network, now with input dimensionality $2$ and output $4$. The number of neurons for the hidden layers was the same as above for the forward model. 

\subsection{Meta-Material}
The forward model architecture consists of a \textit{convolutional network} appended to a fully-connected network with input dimensionality of $8$ and output dimensionality $150$. The remaining layers of the fully-connected network had number of neurons: 1000, 1000, 1000, 1000. The adjacent convolutional network had layers defined with the following (number of filters, kernel size, stride size) definition: (4, 8, 2); (4, 5, 1); (4, 5, 1). 

Each backward model for MMN had a convolutional network followed by a fully-connected network. The convolutional architecture was the inverse of that of the forward model above, and the fully-connected network had input dimensionality $150$ and output dimensionality $8$, with hidden layer sizes: 250, 250, 250, 250 neurons.

\subsection{Shell}
The forward model architecture consists of a fully-connected network with input dimensionality of $8$ and output dimensionality $201$. There were 15 hidden layers of 1700 neurons each.

Each backward model for MMN also consisted of a fully-connected network, now with input dimensionality $201$ and output $8$. There were 14 hidden layers of 2000 neurons each. 

\section{Forward Model Generation: Further Sampling Details}

In this section of our supplemental information, we further specify the distributions from which we sampled points used to generate new training data from a (trained) forward model. Such a method is the second innovation that we outline in our main paper, and our experiments demonstrate that using such generation dramatically improvements the results of our method. For each of the different benchmarks, the distributions we sampled from differed, and this is to correspond with the distributions used to generate the original training data for each benchmark. Recall, as described in our main paper, that the strategy here is to sample points $\tilde{x}$ that we then ``push" through a trained forward model, $\hat{f}$, to generate new points, $\tilde{y}$, with which to train the backward models that compose a MMN. 

\subsection{Sine Wave}
Since the original training data has points $x$ sampled from a uniform distribution on the interval $[-1, 1]$, we also use this distribution from which to sample each dimension for $x$ before pushing through our trained forward model, $\hat{f}$, to generated new (synthetic) training points $\tilde{y}$ with which to train our backward models:
\begin{align*}
    \tilde{x}_i & \sim \mathrm{Uniform}(-1, 1), \ i \in {1, 2} 
\end{align*}

\subsection{Robotic Arm}
Again, we sample points $\tilde{x}$ from distributions that reflect the distributions used to generate the original data:

\begin{equation*}
    \tilde{x}_i \sim \mathcal{N}(0, \sigma_i^2), \ \boldsymbol{\sigma}^2 = \Big[\frac{1}{16}, \frac{1}{4}, \frac{1}{4}, \frac{1}{4}\Big]
\end{equation*}

\subsection{Meta-Material}
We sample points $\tilde{x}$ from the following distributions for the Meta-Material inverse problem:

\begin{align*}
    \tilde{x}_i & \sim \mathrm{Uniform}(-1, 1.273), \ i \in \{1, 2, 3, 4\} \\
    \tilde{x}_i & \sim \mathrm{Uniform}(-1, 1), \ i \in \{5, 6, 7, 8\}
\end{align*}

\subsection{Shell}
For the Shell inverse problem, to sample $\tilde{x}$, we use the same scheme used to sample the original $x$. This involves first sampling, and then transforming the data as outlined below. We have discerned this scheme from the Python implementation for generation of Shell data from \cite{ren2022inverse}.  To sample the points $\tilde{x}$, we perform the following:

\begin{align*}
    \tilde{x}^{'} & \sim \mathrm{Discrete Uniform}(30, 70) \\
    \tilde{x} & = \frac{\tilde{x}^{'} - 50}{20}
\end{align*}
\end{document}